\documentclass[10pt,twocolumn,letterpaper]{article}

\usepackage{cvpr}
\usepackage{times}
\usepackage{epsfig}
\usepackage{graphicx}
\usepackage{amsmath}
\usepackage{amssymb}
\usepackage{color}
\usepackage{pifont}
\usepackage{booktabs}
\usepackage{multirow}
\usepackage{enumitem}
\newcommand{\tabincell}[2]{\begin{tabular}{@{}#1@{}}#2\end{tabular}}

\usepackage[pagebackref=true,breaklinks=true,colorlinks,bookmarks=false]{hyperref}

\begin{document}
\title{Dehaze-then-Splat: Generative Dehazing with Physics-Informed\\3D Gaussian Splatting for Smoke-Free Novel View Synthesis}

\author{
Boss Chen\quad Hanqing Wang \\[4pt]
{\small The Hong Kong University of Science and Technology (Guangzhou)} \\[2pt]
{ \quad hwang201@connect.hkust-gz.edu.cn}
}

\maketitle

\begin{abstract}
We present Dehaze-then-Splat, a two-stage pipeline for multi-view smoke removal and novel view synthesis developed for Track~2 of the NTIRE 2026 3D Restoration and Reconstruction Challenge~\cite{ntire2026_3drr}.
In the first stage, we produce pseudo-clean training images via per-frame generative dehazing using Nano Banana Pro, followed by brightness normalization.
In the second stage, we train 3D Gaussian Splatting (3DGS) with physics-informed auxiliary losses---depth supervision via Pearson correlation with pseudo-depth, dark channel prior regularization, and dual-source gradient matching---that compensate for cross-view inconsistencies inherent in frame-wise generative processing.
We identify a fundamental tension in dehaze-then-reconstruct pipelines: per-image restoration quality does not guarantee multi-view consistency, and such inconsistency manifests as blurred renders and structural instability in downstream 3D reconstruction.
Our analysis shows that MCMC-based densification with early stopping, combined with depth and haze-suppression priors, effectively mitigates these artifacts.
On the Akikaze validation scene, our pipeline achieves 20.98\,dB PSNR and 0.683 SSIM for novel view synthesis, a +1.50\,dB improvement over the unregularized baseline.
\end{abstract}

\section{Introduction}
\label{sec:intro}

Multi-view smoke and haze removal poses a joint challenge: restoring visual quality in each degraded view while maintaining cross-view photometric and geometric consistency for downstream 3D reconstruction.
This paper describes our solution for Track~2 of the NTIRE 2026 3D Restoration and Reconstruction (3DRR) Challenge~\cite{ntire2026_3drr,chang2026training, ge2026dual, chang2026beyond, ge2026clip,zheng20263d,liu2026elog,fu2026smokegs,cao2026gensmoke,zhu2026naka,guo2026reliability}, which requires removing physically captured smoke from multi-view images and synthesizing novel views of the clean scene.
The challenge is built on the RealX3D benchmark~\cite{realx3d}, a physically-degraded 3D dataset featuring real-world smoke and low-light conditions for multi-view restoration and reconstruction evaluation.

Our key insight is that \emph{2D dehazing quality sets the performance ceiling for downstream 3D reconstruction}---but high per-image quality alone is insufficient.
State-of-the-art generative dehazing models such as Nano Banana Pro\footnote{Nano Banana Pro refers to the image generation capability of the Gemini API (\texttt{gemini-3-pro-image-preview})~\cite{nanobanana}.} produce visually compelling per-frame results, yet process each view independently without cross-view conditioning.
This frame-wise independence introduces stochastic variations in color, texture, and hallucinated detail that, when used as training data for 3D Gaussian Splatting (3DGS)~\cite{kerbl3dgs}, cause blurred renders, texture drift, and structural instability.

Motivated by this analysis, we propose \textbf{Dehaze-then-Splat}, a two-stage pipeline that combines generative dehazing with physics-informed 3DGS training.
Stage~1 produces pseudo-clean training images via Nano Banana Pro with brightness normalization.
Stage~2 trains 3DGS with auxiliary losses---depth supervision, dark channel prior (DCP) regularization, and dual-source gradient matching---that serve as implicit multi-view consistency regularizers, compensating for the per-frame variations introduced in Stage~1.
On the Akikaze validation scene, our pipeline achieves \textbf{20.98\,dB} PSNR with 0.683 SSIM, a +1.50\,dB improvement over the unregularized baseline.

\begin{figure*}[t]
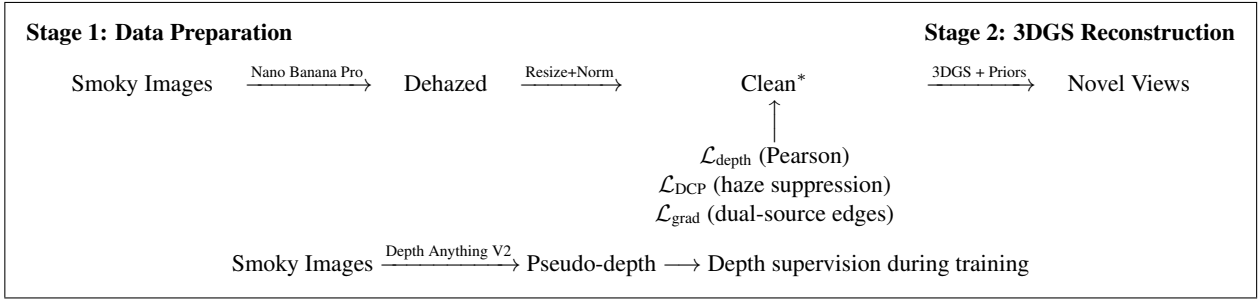

\centering
\setlength{\fboxsep}{8pt}
\fbox{\parbox{0.92\textwidth}{\centering\small
\textbf{Stage 1: Data Preparation} \hfill \textbf{Stage 2: 3DGS Reconstruction} \\[6pt]
\begin{tabular}{ccccccc}
Smoky Images & $\xrightarrow{\text{Nano Banana Pro}}$ & Dehazed & $\xrightarrow{\text{Resize+Norm}}$ & Clean$^*$ & $\xrightarrow{\text{3DGS + Priors}}$ & Novel Views \\
& & & & $\Big\uparrow$ & & \\
& & & & \tabincell{c}{$\mathcal{L}_\text{depth}$ (Pearson) \\ $\mathcal{L}_\text{DCP}$ (haze suppression) \\ $\mathcal{L}_\text{grad}$ (dual-source edges)} & & \\
\end{tabular} \\[4pt]
Smoky Images $\xrightarrow{\text{Depth Anything V2}}$ Pseudo-depth $\longrightarrow$ Depth supervision during training
}}
\caption{Overall pipeline of our Dehaze-then-Splat method. Stage~1 produces pseudo-clean training images via generative dehazing and brightness normalization. Stage~2 trains 3DGS with physics-informed auxiliary losses (depth, DCP, gradient) and MCMC densification with early stopping.}
\label{fig:pipeline}
\end{figure*}

Figure~\ref{fig:pipeline} illustrates the overall architecture.

\section{Method}
\label{sec:method}

\subsection{Generative Dehazing with Nano Banana Pro}
\label{sec:dehazing}

Since 2D dehazing quality directly limits the achievable NVS quality, we prioritize dehazing fidelity.
We employ Nano Banana Pro (\texttt{gemini-3-pro-image-preview}) through the Gemini API~\cite{nanobanana} for per-frame smoke removal.
Each smoky training image is sent with a structured prompt requesting comprehensive smoke removal while preserving scene composition, geometry, and photorealistic appearance.

Nano Banana Pro demonstrates remarkable per-image dehazing capability, producing visually compelling smoke-free outputs that preserve scene geometry and photorealistic appearance.
As shown in Table~\ref{tab:dehaze2d}, it achieves 20.07\,dB per-frame PSNR against ground truth after brightness normalization, substantially outperforming both classical methods such as DCP~\cite{dcp} (11.9\,dB) and learning-based alternatives such as MB-TaylorFormer~\cite{mbtaylor} (17.00\,dB).
The generative nature of the model enables it to hallucinate plausible scene content in heavily occluded regions where discriminative methods produce artifacts.

\noindent\textbf{Prompt engineering.}
A structured 6-point prompt specifying explicit removal targets and preservation constraints outperforms a simple 4-sentence prompt by +0.44\,dB on average.
Scene-specific prompts and few-shot examples proved counterproductive, as additional input complexity amplifies output variance.

\noindent\textbf{Resolution alignment.}
Nano Banana Pro outputs images at approximately $1696{\times}2528$, which does not match the camera intrinsics.
We resize all outputs to the exact resolution in \texttt{transforms\_train.json} to ensure pixel-accurate alignment with camera parameters.

\subsection{Brightness Normalization}
\label{sec:norm}

Frame-wise processing introduces inter-frame brightness inconsistency (max shift of 0.12 between adjacent frames).
Without correction, 3DGS averages over conflicting color signals, degrading both PSNR and visual sharpness.

We apply per-channel mean/std normalization to align each frame's brightness distribution.
When ground-truth clean images are available (Akikaze), we normalize to the GT statistics (+0.8\,dB).
For scenes without GT, we normalize to the median statistics across all frames (self-normalization), reducing the max brightness jump from 0.12 to 0.001.

Note that brightness normalization addresses only global (zero-order) photometric inconsistency.
Local variations in texture detail, hallucinated content, and color rendition across views persist after normalization, as we discuss in the following subsection.

\subsection{The Multi-View Consistency Gap}
\label{sec:consistency}

\begin{figure*}[t]
\centering
\includegraphics[width=\textwidth]{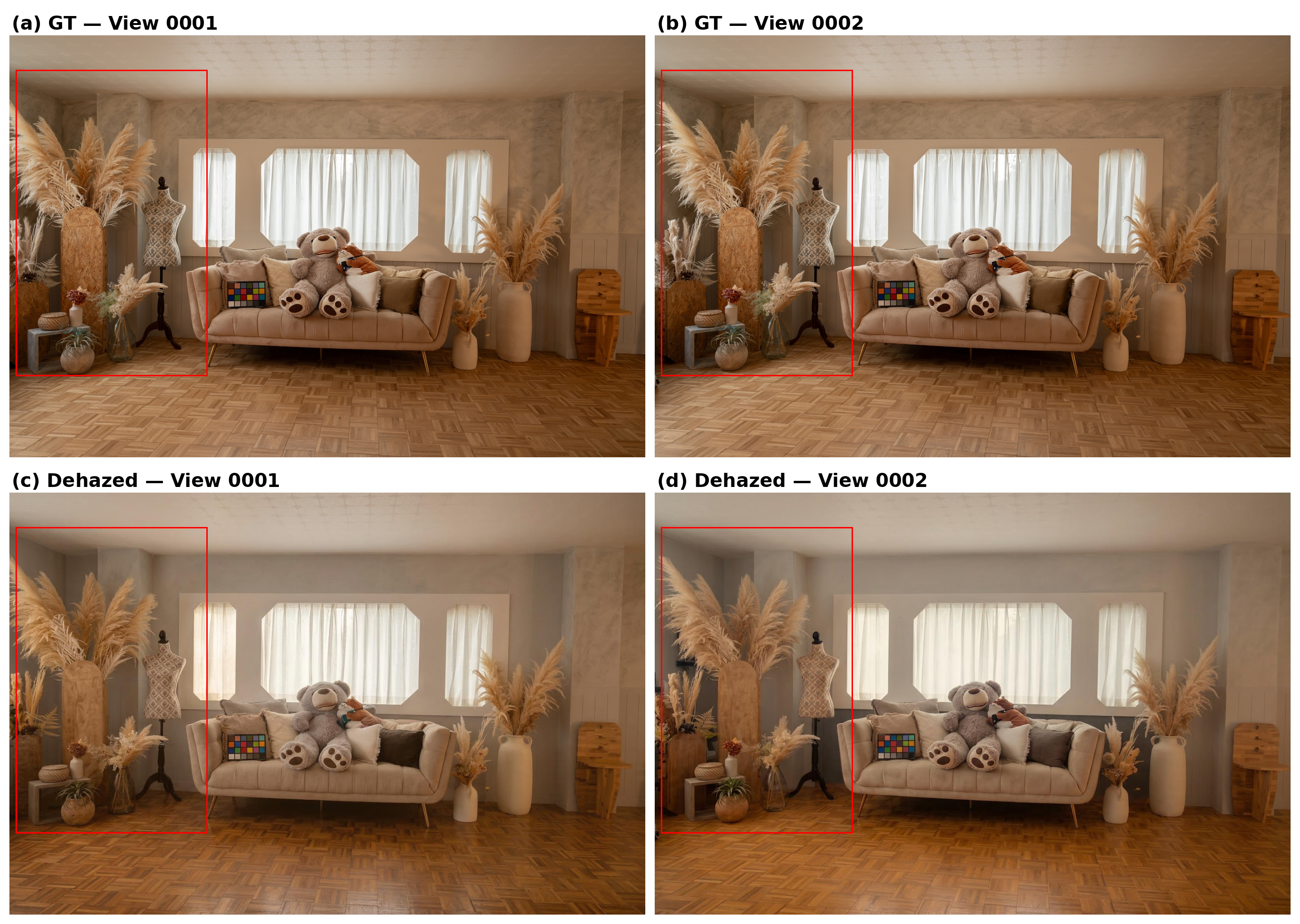}
\caption{Per-frame dehazing quality versus multi-view consistency on Akikaze. (a,\,b)~Ground-truth clean images exhibit consistent color and texture across adjacent views. (c,\,d)~Nano Banana Pro outputs, despite global brightness normalization, show visible differences in color rendition and texture detail. Red boxes highlight the plant region where color shifts are most apparent. Such cross-view inconsistencies degrade downstream 3DGS reconstruction.}
\label{fig:consistency}
\end{figure*}

As a single-image generative model, Nano Banana Pro processes each view independently without access to cross-view correspondences or scene-level priors.
This frame-wise independence introduces stochastic variations in color rendition, texture detail, and hallucinated content across views depicting the same 3D region.
We observe inter-frame brightness standard deviations of up to 0.12 (Table~\ref{tab:dehaze2d}), and---more critically---local texture inconsistencies that persist even after global brightness normalization.
While each individual output may be visually satisfactory, the collection of outputs does not form a multi-view-consistent set suitable for direct 3D reconstruction (Figure~\ref{fig:consistency}).

When trained on such view-inconsistent pseudo-clean images, 3D Gaussian Splatting must reconcile conflicting photometric signals from different viewpoints observing the same scene region.
The optimization responds by broadening Gaussian kernels to average over the discrepancies, resulting in blurred novel-view renders and loss of fine detail.
In more severe cases, the inconsistency manifests as texture flickering across viewpoints, geometric drift in low-texture regions, and floater artifacts where the model allocates additional Gaussians to fit per-view noise rather than genuine scene structure.
This effect is corroborated by our checkpoint analysis (Section~\ref{sec:ablation}): optimal NVS quality occurs at early training steps (step 2000, ${\sim}$104k Gaussians), before the model has sufficient capacity to memorize per-view artifacts.

This observation motivates the physics-informed auxiliary losses described below, which serve as implicit multi-view consistency regularizers.

\subsection{3DGS Training with Physics-Informed Priors}
\label{sec:losses}

To mitigate the cross-view inconsistencies introduced by frame-wise dehazing, we augment standard 3DGS photometric reconstruction with physics-informed auxiliary losses that anchor geometry and suppress residual degradation artifacts.
We train a 3DGS model~\cite{kerbl3dgs} using the gsplat library~\cite{gsplat} with the following loss:
\begin{equation}
\mathcal{L} = (1{-}\lambda_s)\mathcal{L}_1 + \lambda_s \mathcal{L}_\text{SSIM} + \lambda_d \mathcal{L}_\text{DCP} + \lambda_p \mathcal{L}_\text{depth} + \lambda_g \mathcal{L}_\text{grad}
\label{eq:loss}
\end{equation}
where $\lambda_s{=}0.2$, $\lambda_d{=}0.01$, $\lambda_p{=}0.1$, $\lambda_g{=}0.1$.

\noindent\textbf{Depth supervision via Pearson correlation} (most impactful, +0.79\,dB)\textbf{.}
We generate pseudo-depth maps from the original smoky images using Depth Anything V2 (ViT-L)~\cite{depthanything}, which demonstrates strong robustness to smoke degradation.
Since pseudo-depth lacks metric scale, we use a scale-invariant weighted Pearson correlation loss between rendered expected depth $d_r$ and pseudo inverse-depth $d_p$:
\begin{equation}
\mathcal{L}_\text{depth} = 1 + \frac{\sum_i w_i (d_r^i - \bar{d}_r)(d_p^i - \bar{d}_p)}{\sqrt{\sum_i w_i (d_r^i - \bar{d}_r)^2} \cdot \sqrt{\sum_i w_i (d_p^i - \bar{d}_p)^2}}
\end{equation}
where weights $w_i$ are the detached rendering alpha values.

\noindent\textbf{Dark Channel Prior (DCP) regularization.}
The dark channel prior~\cite{dcp} states that clean images have near-zero dark channel values.
We regularize the rendered image $\hat{I}$ so that its dark channel approaches zero:
\begin{equation}
\mathcal{L}_\text{DCP} = \text{mean}\!\Big(\min_{c \in \{R,G,B\}} \big(\text{MinPool}_{k}(\hat{I}_c)\big)\Big)
\end{equation}
where $k{=}15$ is the patch size.

\noindent\textbf{Dual-source gradient loss.}
We supervise edge structure using Sobel gradient matching against a secondary structural reference (MB-TaylorFormer~\cite{mbtaylor} output), after brightness normalization to match the primary source:
\begin{equation}
\mathcal{L}_\text{grad} = \|\nabla \hat{I} - \nabla \text{Norm}(I_\text{struct}, I_\text{nano})\|_1
\end{equation}

\noindent\textbf{MCMC densification strategy.}
We adopt the MCMCStrategy from gsplat~\cite{gsplatmcmc}, which manages Gaussian density through stochastic noise injection and relocation rather than manual split/clone/prune heuristics.
A controlled 2$\times$2 experiment confirms that MCMC provides +0.47\,dB PSNR over DefaultStrategy at matched stopping points.

\noindent\textbf{Early densification stopping.}
Late densification is the primary cause of floater artifacts.
Setting \texttt{DENSIFY\_STOP\_STEP=3000} locks the Gaussian count early (typically 104k--161k), preventing overfitting and floaters in the few-view regime.

\subsection{Alternative: End-to-End Scattering Decomposition}
\label{sec:routeb}

Inspired by DehazeNeRF~\cite{dehazenerf} and DehazeGS~\cite{dehazegs}, we also explored an end-to-end approach integrating the Koschmieder atmospheric scattering model into 3DGS:
$I(x) = J(x) \cdot t(x) + A \cdot (1 - t(x))$, where $t(x) = e^{-\beta \cdot d(x)}$.
Without sufficient geometric constraints, the model collapses to $\beta \to 0$ ($t \to 1$, $J \approx I$), achieving only 10.28\,dB.
This collapse is complementary evidence for our design: just as per-frame 2D dehazing achieves high per-image quality but lacks multi-view consistency, end-to-end 3D dehazing preserves consistency by construction but fails to produce adequate restoration quality without strong scene priors.
The failure of both extremes motivated our hybrid dehaze-then-reconstruct design with physics-informed regularization.

\subsection{Implementation Details}
\label{sec:impl}

\noindent\textbf{Common settings.}
All scenes use: SH degree 3 (progressively increased 0$\to$3), scene scale 2.0, learning rates $\text{lr}_\text{means}{=}1.6{\times}10^{-4}$, $\text{lr}_\text{SH0}{=}2.5{\times}10^{-3}$, total 20k training steps with validation every 1k steps.
\texttt{GAMMA} must be set to 1.0 (the gsplat default 0.5 causes $\sim$5\,dB loss).

\noindent\textbf{Per-scene adaptation.}
Table~\ref{tab:perscene} summarizes scene-specific settings.
The dark scene Hinoki requires reduced initialization (20k vs.\ 50k) and black background to prevent Gaussian explosion.

\begin{table}[t]
\centering
\caption{Per-scene configuration.}
\label{tab:perscene}
\small
\begin{tabular}{@{}lccc@{}}
\toprule
Scene & Init Points & BG Color & Notes \\
\midrule
Akikaze & 50k & 255 & Validation (has GT) \\
Futaba & 50k & 255 & Bright outdoor \\
Koharu & 50k & 255 & High brightness var. \\
Midori & 50k & 255 & Low contrast \\
Hinoki & 20k & 0 & Dark, dense smoke \\
Natsume & 50k & 255 & Test scene \\
Shirohana & 50k & 255 & Test scene \\
Tsubaki & 50k & 255 & Test scene \\
\bottomrule
\end{tabular}
\end{table}

\noindent\textbf{MCMC configuration.}
CAP\_MAX=500k, NOISE\_LR=$5{\times}10^5$ (decayed to 0 after step 8000).
Densification runs from step 500 to \texttt{DENSIFY\_STOP\_STEP} (3000).

\noindent\textbf{Checkpoint selection.}
We save checkpoints every 1000 steps and select the one maximizing PSNR on validation views.
The optimal checkpoint typically appears at step 2000--4000, where the Gaussian count is still low ($\sim$104--161k).
This early optimality is a direct consequence of the multi-view consistency gap (Section~\ref{sec:consistency}): at low Gaussian counts, the model lacks the capacity to overfit to per-view dehazing artifacts and instead learns a view-averaged representation that generalizes to novel viewpoints.

\subsection{Ablation Studies}
\label{sec:ablation}

All ablations are on the Akikaze validation scene (25 training views, 4 test views with GT).

\noindent\textbf{Component-wise ablation.}
Table~\ref{tab:ablation} shows the contribution of each component.

\begin{table}[t]
\centering
\caption{Ablation on Akikaze NVS (4 test views). Best-PSNR checkpoint per row. $\lambda_\text{DCP}{=}0.01$, $\lambda_\text{depth}{=}0.1$, $\lambda_\text{grad}{=}0.1$.}
\label{tab:ablation}
\small
\begin{tabular}{@{}lcccc@{}}
\toprule
Configuration & PSNR$\uparrow$ & SSIM$\uparrow$ & LPIPS$\downarrow$ & Step \\
\midrule
Baseline (Default, STOP=5k) & 19.48 & 0.632 & 0.432 & 20k \\
+DCP+Depth & 20.27 & 0.644 & \textbf{0.396} & 20k \\
+MCMC (STOP=15k) & 20.53 & 0.651 & 0.450 & 20k \\
+MCMC+DCP+Depth (STOP=15k) & 20.57 & 0.655 & 0.450 & 20k \\
+MCMC+DCP+Depth (STOP=3k) & 20.92 & 0.680 & 0.598 & 2k \\
+Dual-source Gradient & \textbf{20.98} & \textbf{0.683} & 0.600 & 2k \\
\bottomrule
\end{tabular}
\end{table}

\begin{figure*}[t]
\centering
\includegraphics[width=\textwidth]{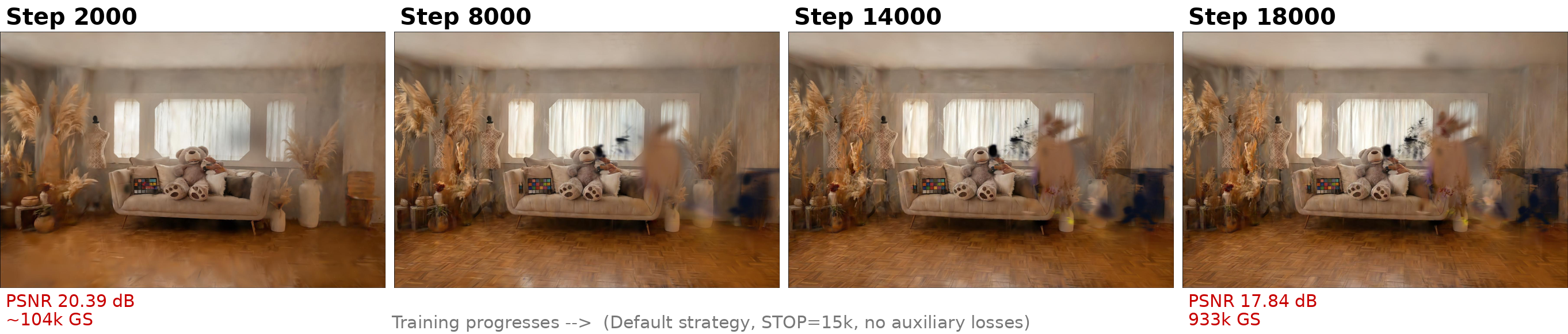}
\caption{Training progression without early stopping (DefaultStrategy, STOP=15k, no auxiliary losses). Novel-view quality peaks at early steps and degrades catastrophically with continued training. By step 18k, Gaussians have exploded from 104k to 933k, producing severe floater artifacts, color bleeding, and structural collapse.}
\label{fig:progression}
\end{figure*}

Key findings: (1)~Depth supervision is the most impactful auxiliary loss (+0.79\,dB), anchoring geometry despite per-view texture variations.
(2)~MCMC outperforms DefaultStrategy (+0.47\,dB at matched STOP).
(3)~Early stopping (STOP=3k) yields +0.35\,dB over STOP=15k by keeping the Gaussian budget tight, limiting the model's capacity to memorize per-view inconsistencies (Section~\ref{sec:consistency}).
(4)~DCP regularization primarily improves perceptual quality by suppressing residual haze.
The consistently early optimal checkpoint (step 2000, ${\sim}$104k Gaussians) across all configurations further supports our analysis: extended training with view-inconsistent data leads to overfitting rather than refinement (Figure~\ref{fig:progression}).

\noindent\textbf{MCMC $\times$ STOP decoupling.}
The initial ablation confounded MCMC strategy with STOP timing.
A clean 2$\times$2 factorial experiment (Table~\ref{tab:decouple}) without auxiliary losses confirms MCMC's genuine +0.47\,dB contribution.

\begin{table}[t]
\centering
\caption{Decoupling MCMC vs.\ STOP (no auxiliary losses, best-PSNR checkpoint). $^\dagger$Catastrophic degradation: PSNR drops to 17.84, Gaussians explode to 933k by step 18k.}
\label{tab:decouple}
\small
\begin{tabular}{@{}lcccc@{}}
\toprule
 & \multicolumn{2}{c}{STOP=5k} & \multicolumn{2}{c}{STOP=15k} \\
\cmidrule(lr){2-3} \cmidrule(lr){4-5}
 & PSNR & SSIM & PSNR & SSIM \\
\midrule
Default & 20.39 & 0.667 & 19.99$^\dagger$ & 0.667$^\dagger$ \\
MCMC & \textbf{20.86} & \textbf{0.677} & 20.58 & 0.674 \\
\bottomrule
\end{tabular}
\end{table}

\begin{figure*}[!t]
\centering
\includegraphics[width=\textwidth]{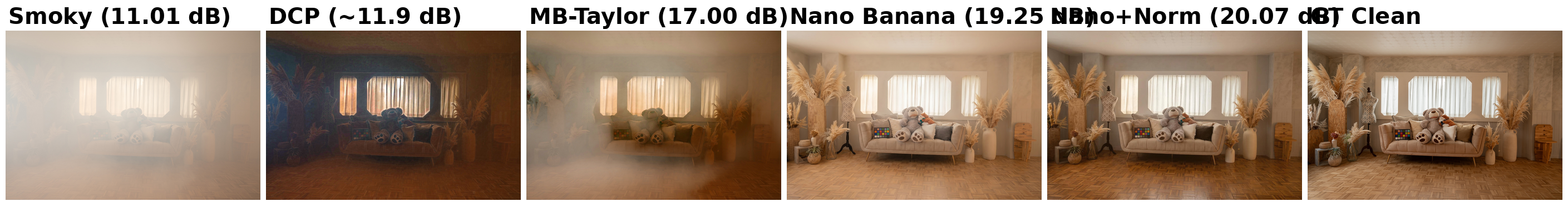}
\caption{Comparison of 2D dehazing methods on Akikaze (view 0001). Nano Banana Pro with GT normalization achieves the highest per-frame PSNR (20.07\,dB), substantially outperforming MB-TaylorFormer (17.00\,dB) and classical DCP ($\sim$11.9\,dB).}
\label{fig:dehaze2d}
\end{figure*}

\noindent\textbf{2D dehazing model comparison.}
Table~\ref{tab:dehaze2d} and Figure~\ref{fig:dehaze2d} show that Nano Banana Pro with GT normalization achieves the best per-frame PSNR (20.07\,dB), closely predicting the NVS ceiling.
Our best NVS result (20.98\,dB) modestly exceeds this because multi-view 3DGS averaging compensates for per-frame noise.

\begin{table}[t]
\centering
\caption{2D dehazing quality on Akikaze (per-frame PSNR vs.\ GT, averaged over 25 training views).}
\label{tab:dehaze2d}
\small
\begin{tabular}{@{}lcc@{}}
\toprule
Method & PSNR & Brightness Std$\downarrow$ \\
\midrule
Nano Banana Pro + GT norm & \textbf{20.07} & 0.012 \\
Nano Banana Pro (raw) & 19.25 & 0.120 \\
Nano Banana 2 & 18.72 & 0.094 \\
MB-TaylorFormer-L & 17.00 & \textbf{0.048} \\
DCP (classical) & $\sim$11.9 & --- \\
Original smoky & 11.01 & --- \\
\bottomrule
\end{tabular}
\end{table}

\subsection{Final Results}
\label{sec:results}

\begin{figure*}[t]
\centering
\includegraphics[width=\textwidth]{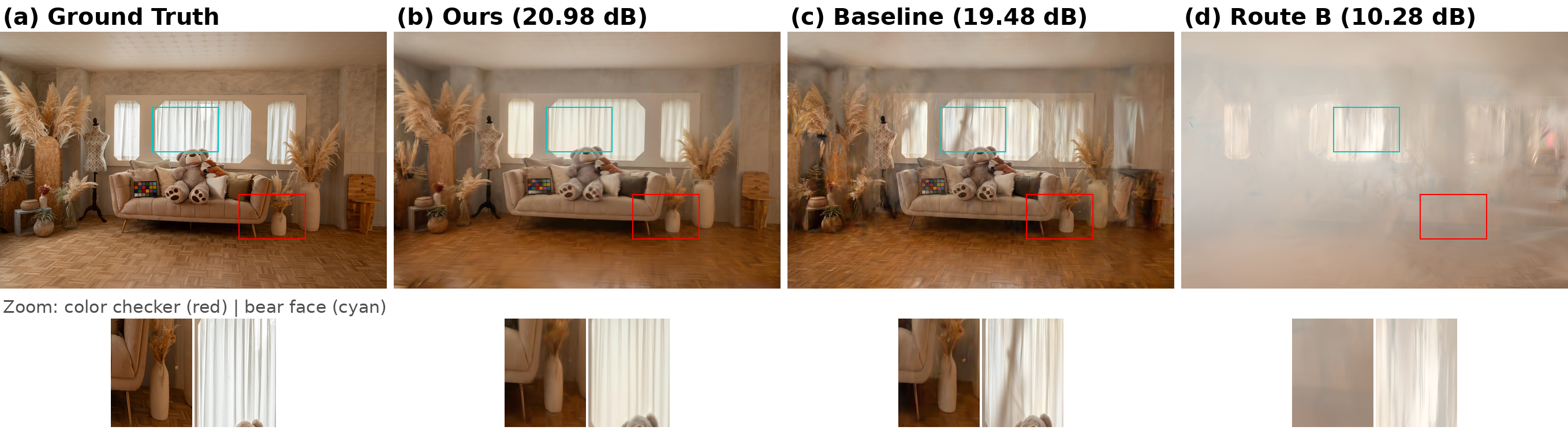}
\caption{Qualitative comparison of novel view synthesis on Akikaze (test view 0026). (a)~Ground truth. (b)~Our full pipeline produces sharp renders with accurate color. (c)~Baseline without auxiliary losses shows moderate blur and color shift. (d)~End-to-end scattering decomposition (Route~B) fails to separate smoke from scene content. Bottom row: zoom crops of color checker (red box) and bear face (cyan box) regions.}
\label{fig:comparison}
\end{figure*}

Table~\ref{tab:final} summarizes our results on the Akikaze validation scene.
The total improvement from baseline to our full pipeline is \textbf{+1.50\,dB PSNR} and \textbf{+0.051 SSIM}.
All eight competition scenes were trained with the STOP=3k, $\lambda_\text{depth}{=}0.1$ configuration and submitted using step-2000 checkpoints.
Figure~\ref{fig:comparison} shows qualitative results on test view 0026.

\begin{table}[t]
\centering
\caption{Final NVS results on Akikaze (4 test views).}
\label{tab:final}
\small
\begin{tabular}{@{}lcccc@{}}
\toprule
Configuration & PSNR & SSIM & LPIPS & Step \\
\midrule
Full pipeline (all losses) & \textbf{20.98} & \textbf{0.683} & 0.600 & 2k \\
w/o gradient loss & 20.92 & 0.680 & 0.598 & 2k \\
Baseline (no aux.\ losses) & 19.48 & 0.632 & 0.432 & 20k \\
Route B (Dehaze3DGS) & 10.28 & 0.573 & 0.780 & 20k \\
\bottomrule
\end{tabular}
\end{table}

\section{Discussion}
\label{sec:discussion}

Our results highlight a fundamental tension in dehaze-then-reconstruct pipelines: methods that excel at per-image restoration do not necessarily produce multi-view-consistent outputs, yet consistency is a prerequisite for high-fidelity 3D reconstruction.
Nano Banana Pro achieves state-of-the-art single-image dehazing quality (20.07\,dB per-frame PSNR), but the downstream 3DGS must compensate for its view-to-view variations through auxiliary losses and careful capacity control.

This tension also explains the failure of the opposite extreme: our end-to-end scattering decomposition (Section~\ref{sec:routeb}) maintains multi-view consistency by construction but collapses without adequate restoration priors.
Neither purely 2D nor purely 3D approaches are sufficient; effective solutions must balance restoration quality with cross-view coherence.

We believe future work integrating cross-view consistency constraints into the dehazing stage---whether through video-based generative models, multi-view attention mechanisms, or test-time consistency optimization---could substantially narrow this gap.
Additionally, adaptive capacity scheduling that allocates more Gaussians only after cross-view consistency is established may offer further improvements.

\section{Conclusion}

We presented Dehaze-then-Splat, a two-stage pipeline for multi-view smoke removal and novel view synthesis.
Our analysis reveals a fundamental tension between per-image dehazing quality and multi-view consistency: generative models such as Nano Banana Pro produce compelling individual frames but introduce cross-view variations that degrade downstream 3D reconstruction.
We address this through physics-informed auxiliary losses and early densification stopping, achieving 20.98\,dB PSNR on the Akikaze validation scene.
Future work should explore consistency-aware dehazing, either through multi-view generative models or test-time consistency optimization, to bridge the gap between 2D restoration and 3D reconstruction.

Code is available at: \url{https://github.com/chen-yu-chao/3DRR_codebase}.

{\small
\bibliographystyle{ieeetr}
\bibliography{main}

@article{realx3d,
  title={{RealX3D}: A Physically-Degraded {3D} Benchmark for Multi-view Visual Restoration and Reconstruction},
  author={Liu, Shuhong and Bao, Chenyu and Cui, Ziteng and Liu, Yun and Chu, Xuangeng and Gu, Lin and Conde, Marcos V. and Umagami, Ryo and Hashimoto, Tomohiro and Hu, Zijian and Xu, Tianhan and Gan, Yuan and Kurose, Yusuke and Harada, Tatsuya},
  journal={arXiv preprint arXiv:2512.23437},
  year={2025},
  url={https://arxiv.org/abs/2512.23437}
}

@article{kerbl3dgs,
  title={{3D} Gaussian Splatting for Real-Time Radiance Field Rendering},
  author={Kerbl, Bernhard and Kopanas, Georgios and Leimk{\"u}hler, Thomas and Drettakis, George},
  journal={ACM Transactions on Graphics},
  volume={42},
  number={4},
  pages={139:1--139:14},
  year={2023},
  doi={10.1145/3592433},
  url={https://doi.org/10.1145/3592433}
}

@article{gsplat,
  title={{gsplat}: An Open-Source Library for Gaussian Splatting},
  author={Ye, Vickie and Turkulainen, Matias and Gao, Jiawei and Levis, Zachary and Loop, Charles and Yumer, Ersin and Tenenbaum, Josh and Slack, Dylan and Su, Hsiao-Yu and Tang, Jiaming and Kanazawa, Angjoo},
  journal={Journal of Machine Learning Research},
  volume={26},
  number={34},
  pages={1--17},
  year={2025},
  url={https://www.jmlr.org/papers/v26/24-1476.html}
}

@inproceedings{gsplatmcmc,
  title={{3D} Gaussian Splatting as Markov Chain Monte Carlo},
  author={Kheradmand, Shakiba and Rebain, Daniel and Sharma, Gopal and Sun, Weiwei and Tseng, Yang-Che and Isack, Hossam and Tagliasacchi, Andrea and Kar, Abhishek and Yi, Kwang Moo},
  booktitle={Advances in Neural Information Processing Systems},
  volume={37},
  year={2024},
  url={https://papers.nips.cc/paper_files/paper/2024/hash/93be245fce00a9bb2333c17ceae4b732-Abstract-Conference.html}
}

@inproceedings{depthanything,
  title={{Depth Anything V2}},
  author={Yang, Lihe and Kang, Bingyi and Huang, Zilong and Zhao, Zhen and Xu, Xiaogang and Feng, Jiashi and Zhao, Hengshuang},
  booktitle={Advances in Neural Information Processing Systems},
  volume={37},
  year={2024},
  url={https://papers.nips.cc/paper_files/paper/2024/hash/26cfdcd8fe6fd75cc53e92963a656c58-Abstract-Conference.html}
}

@inproceedings{dcp,
  title={Single Image Haze Removal Using Dark Channel Prior},
  author={He, Kaiming and Sun, Jian and Tang, Xiaoou},
  booktitle={IEEE Conference on Computer Vision and Pattern Recognition},
  pages={1956--1963},
  year={2009},
  doi={10.1109/CVPR.2009.5206515},
  url={https://doi.org/10.1109/CVPR.2009.5206515}
}

@misc{nanobanana,
  title={{Nano Banana} Image Generation},
  author={{Google AI for Developers}},
  year={2026},
  howpublished={\url{https://ai.google.dev/gemini-api/docs/nanobanana}},
  note={Official Gemini API documentation; accessed March 26, 2026}
}

@inproceedings{mbtaylor,
  title={{MB-TaylorFormer}: Multi-Branch Efficient Transformer Expanded by Taylor Formula for Image Dehazing},
  author={Qiu, Yuwei and Zhang, Kaihao and Wang, Chenxi and Luo, Wenhan and Li, Hongdong and Jin, Zhi},
  booktitle={Proceedings of the IEEE/CVF International Conference on Computer Vision},
  pages={12802--12813},
  year={2023},
  url={https://openaccess.thecvf.com/content/ICCV2023/html/Qiu_MB-TaylorFormer_Multi-Branch_Efficient_Transformer_Expanded_by_Taylor_Formula_for_Image_ICCV_2023_paper.html}
}

@inproceedings{dehazenerf,
  title={{DehazeNeRF}: Multi-image Haze Removal and {3D} Shape Reconstruction Using Neural Radiance Fields},
  author={Chen, Wei-Ting and Wang, Yifan and Kuo, Sy-Yen and Wetzstein, Gordon},
  booktitle={2024 International Conference on 3D Vision},
  pages={247--256},
  year={2024},
  doi={10.1109/3DV62453.2024.00039},
  url={https://doi.org/10.1109/3DV62453.2024.00039}
}

@article{ntire2026_3drr,
  title={{NTIRE} 2026 {3D} Restoration and Reconstruction in Real-world Adverse Conditions: {RealX3D} Challenge Results},
  author={Liu, Shuhong and Bao, Chenyu and Cui, Ziteng and Chu, Xuangeng and Ren, Bin and Gu, Lin and Chen, Xiang and Li, Mingrui and Ma, Long and Conde, Marcos V. and others},
  journal={arXiv preprint arXiv:2604.04135},
  year={2026},
  url={https://arxiv.org/abs/2604.04135}
}

@article{dehazegs,
  title={{DehazeGS}: {3D} Gaussian Splatting for Multi-image Haze Removal},
  author={Ma, Chenjun and Zhao, Jieyu and Chen, Jian},
  journal={IEEE Signal Processing Letters},
  volume={32},
  pages={736--740},
  year={2025},
  doi={10.1109/LSP.2025.3530852},
  url={https://doi.org/10.1109/LSP.2025.3530852}
}

@article{chang2026training,
 	title={Training-Free Model Ensemble for Single-Image Super-Resolution via Strong-Branch Compensation},
 	author={Chang, Gengjia and Ge, Xining and Yuan, Weijun and Li, Zhan and Song, Qiurong and Zhu, Luen and Liu, Shuhong},
 	journal={arXiv preprint arXiv:2604.11564},
 	year={2026}
}

@article{ge2026dual,
title={Dual-Branch Remote Sensing Infrared Image Super-Resolution},
 	author={Ge, Xining and Chang, Gengjia and Yuan, Weijun and Li, Zhan and Chen, Zhanglu and Yao, Boyang and Chen, Yihang and Deng, Yifan and Liu, Shuhong},
 	journal={arXiv preprint arXiv:2604.10112},
 	year={2026}
}

@article{chang2026beyond,
title={Beyond Model Design: Data-Centric Training and Self-Ensemble for Gaussian Color Image Denoising},
author={Chang, Gengjia and Ge, Xining and Yuan, Weijun and Li, Zhan and Song, Qiurong and Zhu, Luen and Liu, Shuhong},
journal={arXiv preprint arXiv:2604.11468},
year={2026}
}

@article{ge2026clip,
title={Clip-guided data augmentation for night-time image dehazing},
 	author={Ge, Xining and Yuan, Weijun and Chang, Gengjia and Li, Xuyang and Liu, Shuhong},
 	journal={arXiv preprint arXiv:2604.05500},
 	year={2026}
}

@article{zheng20263d,
 	title={3D Smoke Scene Reconstruction Guided by Vision Priors from Multimodal Large Language Models},
 	author={Zheng, Xinye and Wang, Fei and Nie, Yiqi and Li, Kun and Chen, Junjie and Zhao, Jiaqi and Wei, Yanyan and Wu, Zhiliang},
 	journal={arXiv preprint arXiv:2604.05687},
 	year={2026}
}

@article{liu2026elog,
 	title={ELoG-GS: Dual-Branch Gaussian Splatting with Luminance-Guided Enhancement for Extreme Low-light 3D Reconstruction},
 	author={Liu, Yuhao and Wang, Dingju and Zheng, Ziyang},
 	journal={arXiv preprint arXiv:2604.12592},
 	year={2026}
}

@article{fu2026smokegs,
 	title={SmokeGS-R: Physics-Guided Pseudo-Clean 3DGS for Real-World Multi-View Smoke Restoration},
 	author={Fu, Xueming and Han, Lixia},
 	journal={arXiv preprint arXiv:2604.05301},
 	year={2026}
}

@article{cao2026gensmoke,
 	title={GenSmoke-GS: A Multi-Stage Method for Novel View Synthesis from Smoke-Degraded Images Using a Generative Model},
 	author={Cao, Qida and Hu, Xinyuan and Shi, Changyue and Ding, Jiajun and Yu, Zhou and Yu, Jun},
 	journal={arXiv preprint arXiv:2604.03039},
 	year={2026}
}

@article{zhu2026naka,
 	title={Naka-GS: A Bionics-inspired Dual-Branch Naka Correction and Progressive Point Pruning for Low-Light 3DGS},
 	author={Zhu, Runyu and Dong, SiXun and Zhang, Zhiqiang and Ye, Qingxia and Xu, Zhihua},
 	journal={arXiv preprint arXiv:2604.11142},
 	year={2026}
}

@article{guo2026reliability,
  	title   = {Reliability-Aware Staged Low-Light Gaussian Splatting},
 	author  = {Guo, Haojie and Xian, Ke},
  	journal = {ResearchGate preprint},
 	year    = {2026}
}
}

\end{document}